# Local Gradient Hexa Pattern: A Descriptor for Face Recognition and Retrieval

Soumendu Chakraborty, Satish Kumar Singh, and Pavan Chakraborty

**Abstract**: Local descriptors used in face recognition are robust in a sense that these descriptors perform well in varying pose, illumination and lighting conditions. Accuracy of these descriptors depends on the precision of mapping the relationship that exists in the local neighborhood of a facial image into microstructures. In this paper a local gradient hexa pattern (LGHP) is proposed that identifies the relationship amongst the reference pixel and its neighboring pixels at different distances across different derivative directions. Discriminative information exists in the local neighborhood as well as in different derivative directions. Proposed descriptor effectively transforms these relationships into binary micropatterns discriminating interclass facial images with optimal precision. Recognition and retrieval performance of the proposed descriptor has been compared with state-of-the-art descriptors namely LDP and LVP over the most challenging and benchmark facial image databases, i.e. Cropped Extended Yale-B, CMU-PIE, color-FERET, and LFW. The proposed descriptor has better recognition as well as retrieval rates compared to state-of-the-art descriptors.

## 1. Introduction

*1.1 Motivation*

Local descriptors are gaining more and more recognition in recent years as these descriptors are capable enough to identify the unique features, which suitably and uniquely describe any image for recognition and retrieval. The original problem was that of the facial region identification and extraction of facial features (eye, nose and mouth) [1-2]. The identification problem has been evolved as a classification problem where an image is classified as a face or non-face. Some of the mathematical models used for classification are based on singular value decomposition [3], neural networks [4], principle component analysis (PCA) [5-6], variations of PCA [7-10], and linear discriminant analysis (LDA) [5-6][11-13]. Rotation invariant feature descriptor, local binary pattern (LBP) and its variants have been proposed for texture classification [14-16]. LBP was later extended to identify the spatial relationships amongst the pixels in the local neighborhood of the facial images [17]. Color feature based descriptor proposed in [18] extracts M color component features to describe a facial image for recognition. Color texture feature extraction proposed in [19] applies Gabor wavelet and LBP on different color channels to extract the facial features. Local directional number pattern [20] computes the edges in an image using compass mass and selects most positive and negative directions as the features. Most of the existing local descriptors tend to confine the local neighborhood to the eight pixels nearest to the reference pixel (current pixel under consideration). These descriptors tend to ignore the discriminating information that exists at different radial widths as well as across higher order derivative spaces.

*1.2 Related work*

More discriminative features that exist in higher order derivative directions were captured by the local derivative pattern (LDP) [21], an improvement over LBP. LDP computes the derivative of a facial image in four directions and encodes the relationship of the reference pixel and its eight neighbors in different derivative directions. Local tetra pattern (LTrP) [22] divides the image in two derivatives along 0° and 90° and encodes the directions of the reference pixel and its eight neighbors computed from these derivatives. Local tetra pattern (LTrP) is an extension of LDP and shows some improvement over LDP with respect to retrieval rate. Local vector pattern (LVP) [27] computes vectors in four unique directions which is similar to the derivatives computed as in case of LDP. Pair wise vectors are encoded using comparative space transform (CST) [27] to generate the micropatterns. LVP is an effective extension over LTrP. Most of these descriptors discard the information that exist at higher radial distances, which adversely affect the retrieval and recognition accuracies. Proposed local gradient hexa pattern (LGHP) encodes the discriminative relationship amongst the reference pixel and its eight neighbors at different distances in four distinct derivative directions using a unique encoding scheme and achieves better recognition and retrieval rate compared to LDP and LVP.

*1.3 Major contribution*

Descriptors that explore higher order derivative space tend to achieve better results under pose, expression, light, and illumination variations. LDP is a local descriptor that explores the higher order derivative space but tend to ignore the relationships amongst pixels at greater radial widths. LDP identifies the relationships within the same derivative space and discards the relationships that exist between the higher order derivative spaces (i.e. inter derivative space relationships). LVP is another descriptor which ignores the relationships amongst pixels at greater radial widths. However, it captures the discriminating information between derivative spaces i.e. between $(0°, 45°), (45°, 90°), (90°, 135°)$ and $(135°, 180°)$.

The proposed descriptor captures discriminating information across different derivative spaces namely $G_{0°,R}^1$, $G_{45°,R}^1$,

$G^1_{90°,R}$, and $G^1_{135°,R}$ at different angular widths 0°, 45°, 90°, and 135° respectively at different radial widths $R$. It also captures discriminating information between derivative spaces i.e. between $(0°, 45°), (0°, 90°), (0°, 135°), (45°, 90°), (45°, 135°)$, and $(90°, 135°)$. Unlike LVP and LDP the proposed descriptor encodes most of the relevant information that exists in the local neighborhood of the reference pixel across different angular widths, radial widths, and higher order derivative spaces including inter derivative space relationships. The result analysis objectively proves that the proposed descriptor encodes additional relevant information, which is ignored by the state of the art descriptors LVP and LDP.

The organization of the rest of the paper is as follows. Section II elaborates the proposed descriptor. Performance measures are defined in section III. The performance of the proposed descriptor has been analyzed and compared with the LVP and LDP in section IV. We conclude in section V.

## 2. Local Gradient Hexa Pattern

LGHP identifies the variations in the original pixels of a facial image in the higher order derivative space. The major difference between LVP and LGHP is the encoding function and the structure of the micropattern.

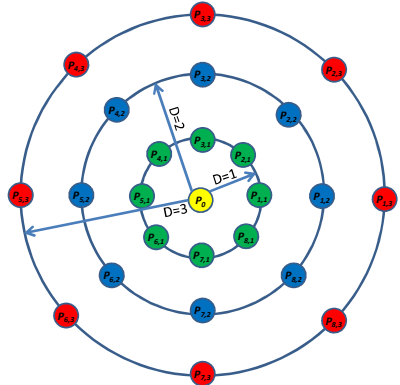

Fig. 1. Template of the local sub-region of an image.

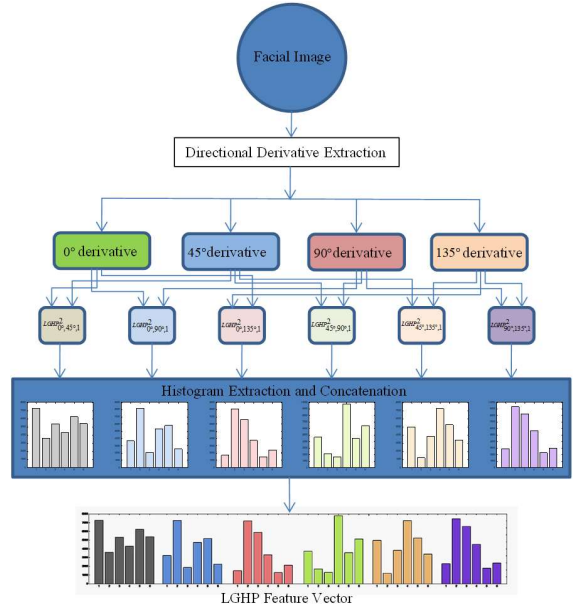

Fig. 2. Flow diagram of the LGHP computation.

Four first order directional derivatives (gradients) $G^1_{\alpha,D}$ of the image $I(P)$ of size $M \times N$ in $\alpha = 0°$, $45°, 90°, and\ 135°$ directions at distance $D$ for an arbitrary reference point $P_0$ are computed as follows

$$G^1_{\alpha,R}(P_0) = I(P_0) - I\left(P_{(\frac{\alpha}{45}+1),D}\right) \quad (1)$$

Fig.1. shows the reference pixel and the eight neighbors at different distances. $P_0$ is the reference pixel and pixel in $0°$ direction at a distance $D = 1$ is represented as $P_{(\frac{0}{45}+1),1}$ that is $P_{1,1}$.

Fig.2 shows the flow diagram of the LGHP computation process. Second order $LGHP^2_{\alpha,\beta,D}(.)$ at a $D$ distance is computed by encoding the pair wise derivatives using encoding function $C(.,.)$ and concatenating these encoded patterns. $LGHP^2_{\alpha,\beta,D}(.)$ is defined as

$$LGHP^2_{\alpha,\beta,D}(P_0) = \left\{C\left(G^1_{\alpha,D}(P_0), G^1_{\beta,D}(P_0)\right), C\left(G^1_{\alpha,D}(P_{1,D}), G^1_{\beta,D}(P_{1,D})\right), \ldots \quad C\left(G^1_{\alpha,D}(P_{8,D}), G^1_{\beta,D}(P_{8,D})\right)\right\} \quad (2)$$

Where $\alpha, \beta = 0°, 45°, 90°, and\ 135°$

The encoding function $C(.,.)$ for any point $P$ in the derivative space is defined as

$$C\left(G^1_{\alpha,D}(P), G^1_{\beta,D}(P)\right) = \begin{cases} 1, & if\ G^1_{\alpha,D}(P) > G^1_{\beta,D}(P) \\ 0, & else \end{cases} \quad (3)$$

Second order LGHP at a distance $D$ is calculated by concatenating $LGHP^2_{\alpha,\beta,D}(P_0)$ for different $(\alpha,\beta)$ pairs as follows

$$LGHP^2_D(P_0) = \{LGHP^2_{\alpha,\beta,D}(P_0)|(\alpha,\beta) = (0°, 45°), (0°, 90°), (0°, 135°), (45°, 90°),$$

$$(45°, 135°), (90°, 135°)\} \quad (4)$$

The proposed LGHP depends on another parameter $R \in [1, N]$ which specifies the upper limit of $D$ over which hexa patterns are computed.

Finally second order LGHP is computed as

$$LGHP_R^2(P_0) = \{LGHP_D^2(P_0) | D = 1, 2, 3 \ldots R\} \quad (5)$$

LGHP encodes six binary patterns of nine bits each at a particular distance for different $(\alpha, \beta)$ pairs. These six patterns are converted to equivalent decimal values to generate six LGHP matrices. Spatial histograms of these six matrices are computed as

$$HLGHP(\alpha, \beta, D) = \{H_{LGHP_{\alpha,\beta,D}} | (\alpha, \beta) = (0°, 45°), (0°, 90°), (0°, 135°), (45°, 90°),$$

$$(45°, 135°), (90°, 135°)\} \quad (6)$$

Where $H_{LGHP_{\alpha,\beta,D}}$ is the histogram extracted from LGHP matrices corresponding to different pairs of $\alpha$ and $\beta$. $L1$ is used to measure the similarity between two histograms as it performs better than other measures on the datasets used in the experiments. Similarity measure $S_{L1}(.,.)$ is defined as

$$S_{L1}(X, Y) = \sum_{i=0}^{q} |x_i - y_i| \quad (7)$$

where $S_{L1}(X, Y)$ is the $L1$ distance computed on two vectors $X = (x_1, \ldots, x_q)$ and $Y = (y_1, \ldots, y_q)$. Nearest one neighbor (1 NN) classifier is used as used in [21] to compute the minimum $L1$ distance between the probe image and the gallery images. As similar regions of the probe and gallery images are effectively identified by 1NN classifier with optimal computational cost [21].

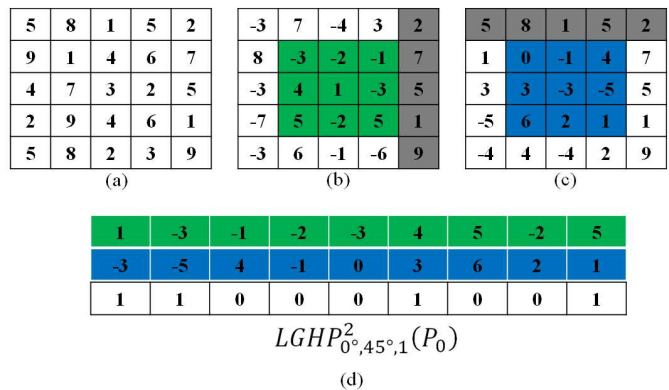

Fig.3. (a) Original sample image, (b) $G_{0°,1}^1(P_0)$ of original sample, (c) $G_{45°,1}^1(P_0)$ of original sample, and (d) Illustration of computation of $LGHP_{\alpha,\beta,D}^2(P_0)$ for $\alpha = 0°$ and $\beta = 45°$

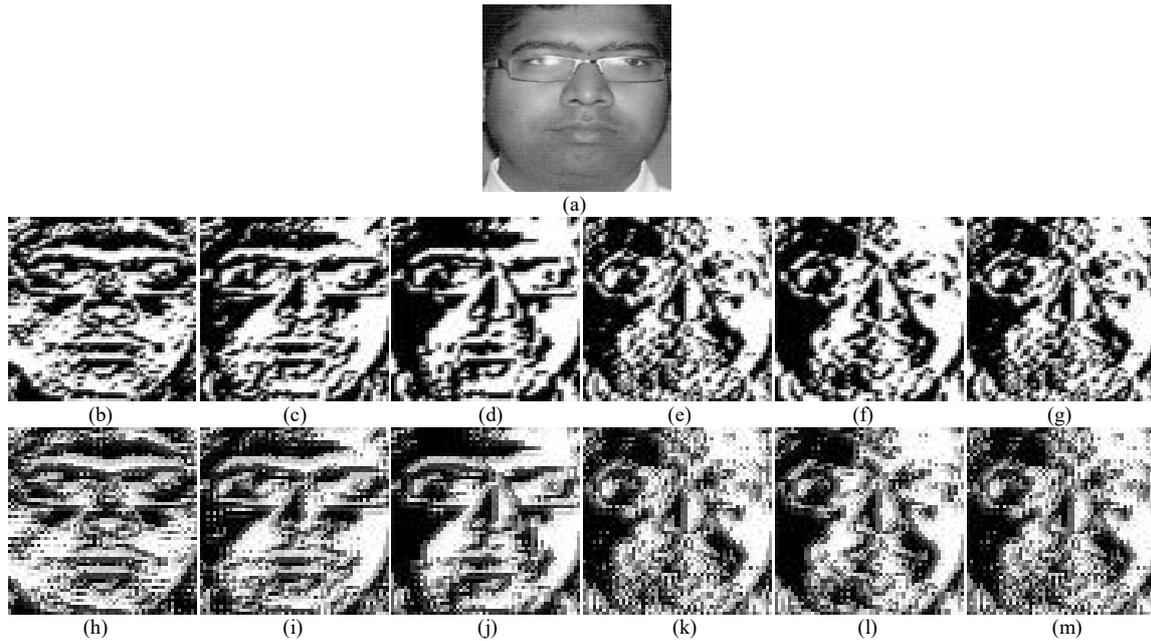

Fig.4. Feature images (FIs): (a) Original, (b) FI at D=1 and direction pair (0°, 45°), (c) FI at D=1 and direction pair (0°, 90°), (d) FI at D=1 and direction pair (0°, 135°), (e) FI at D=1 and direction pair (45°, 90°), (f) FI at D=1 and direction pair (45°, 135°), (g) FI at D=1 and direction pair (90°, 135°), (h) FI at D=2 and direction pair (0°, 45°), (i) FI at D=2 and direction pair (0°, 90°), (j) FI at D=2 and direction pair (0°, 135°), (k) FI at D=2 and direction pair (45°, 90°), (l) FI at D=2 and direction pair (45°, 135°), (m) FI at D=2 and direction pair (90°, 135°)

Fig.3(a) shows a sample image and its derivatives in 0° and 45° directions are shown in Fig.3(b) and Fig.3(c) respectively. Reference pixel and its eight neighbors in 0° and 45° derivative directions are shown in green and blue color respectively. Fig.3(d) shows the computation of the 9 bit binary pattern shown in white row from 0° derivative shown in green row and 45° derivative shown in blue row. Similarly 5 remaining 9 bit patterns for direction pairs (0°, 90°), (0°, 135°), (45°, 90°), (45°, 135°), and (90°, 135°) are computed to generate the hexa pattern at a distance D.

Feature images (FIs) computed using LGHP shown in Fig.4 visually depict the discriminating information captured by the descriptor. As shown in Fig.4 feature images computed at different distances encode useful relationships amongst neighborhood pixels, which help in discriminating interclass facial images. Feature images shown in Fig. 4(b-g) and Fig. 4(h-m) visually demonstrate that while considering the facial features at greater distances the facial features at smaller distance should not be discarded.

The novelty of the proposed descriptor over LVP is in a sense that the proposed descriptor identifies discriminating relationship between the local neighborhoods of the higher order derivatives of the original image with two distance parameters D and R and six different pairs of derivative directions. The novelty in the concept and advantages of the proposed LGHP can be summarized as follows.

1) LDP identifies the discriminating relationship within the directional derivative space at a radius R = 1. LVP extends LDP and explores relationships existing in the inter directional derivative space using directional derivative pairs (0°, 45°), (45°, 90°), (90°, 135°), and (135°, 180°). Even though LVP is computed at different radii, descriptor ignores neighboring pixels at smaller radii while computing it for larger radii (e.g. the descriptor ignores neighboring pixels at R = 1 and R = 2 while computing LVP for R = 3. The proposed LGHP is also an extension of the LDP, which computes relationships existing in directional derivative pairs (0°, 45°), (0°, 90°), (0°, 135°)(45°, 90°), (45°, 135°), and (90°, 135°). LGHP is computed at R = 3 without ignoring the fact that neighboring pixels at R = 1 and R = 2 also contributes to the discriminating information captured using the encoding function.

2) Encoding functions of LDP, LVP, and LGHP are vastly different from each other. LVP ignores the center pixel in the directional derivative space while computing the micropatterns, which is included in the micropatterns computed using LGHP.

3) LGHP is able to extract the features at varying locality of the facial image. As the size of the locality (R) is increased the discriminating power of the proposed descriptor is also increased. This shows that LDP and LVP ignore important relationships while computing the micropatterns.

4) The proposed descriptor works even better compared to LDP and LVP in unconstrained environment as shown in performance analysis.

3. Performance measures

Performance of the proposed LGHP has been analyzed with respect to retrieval and recognition accuracies. Performance measures used to evaluate retrieval accuracy are Average Precision Rate (APR) and Average Retrieval Rate (ARR). Precision is computed as

$$P_r(I_q, n) = \frac{1}{n} \sum_{i=1}^{|DS|} \Delta\left(\omega(I_q), \omega(I_i), \tau(I_q, I_i), n\right) | I_i \neq I_q \quad (8)$$

where $n$ is the number of images retrieved, $I_q$ is the query image, $|DS|$ is the size of the dataset, $\omega(.)$ returns the class of an image and $\tau(I_q, I_i)$ is the rank of the $i^{th}$ image with respect to the query image $I_q$. Image rank is computed using similarity measure $S_{L1}(.,.)$ between the $i^{th}$ image in the dataset and the query image. $\Delta(.)$ is a binary function defined as

$$\Delta(\omega(I_q), \omega(I_i), \tau(I_q, I_i), n) = \begin{cases} 1, & \omega(I_q) = \omega(I_i) \text{ and } \tau(I_q, I_i) \leq n \\ 0, & else \end{cases} \quad (9)$$

Average precision per class is computed as

$$AP(C_i, n) = \frac{1}{|C_i|} \sum_{q=1}^{|C_i|} P_r(I_q, n) \tag{10}$$

where $C_i$ denotes the $i^{th}$ class in the dataset and $|C_i|$ denotes the number of images in the $i^{th}$ class. APR is calculated over the entire dataset of $N_c$ distinct classes as

$$APR(N_c) = \frac{1}{N_c} \sum_{i=1}^{N_c} AP(C_i, n) \tag{11}$$

Recall is defined as

$$R_e(I_q, C_i, n) = \frac{1}{|C_i|} \sum_{i=1}^{|DS|} \Delta\big(\omega(I_q), \omega(I_i), \tau(I_q, I_i), n\big) \mid I_i \neq I_q \tag{12}$$

Average recall per class and ARR over the entire dataset are calculated as

$$AR_e(C_i) = \frac{1}{|C_i|} \sum_{q=1}^{|C_i|} R_e(I_q, C_i) \tag{13}$$

$$ARR(N_c) = \frac{1}{N_c} \sum_{i=1}^{N_c} AR_e(C_i) \tag{14}$$

Recognition rate $\gamma$ is used to measure the recognition accuracy of the proposed LGHP. Recognition rate is defined as

$$\gamma = \left(\frac{1}{|DS|} \sum_{probe}^{|DS|} \sum_{i=1}^{|DS|} \hat{\Delta}\big(\omega(I_{probe}), \omega(I_i), \tau(I_{probe}, I_i)\big)\right) \times 100 \tag{15}$$

$\hat{\Delta}(.)$ computes 2$^{nd}$ best match for the probe image $I_{probe}$ as

$$\hat{\Delta}\big(\omega(I_{probe}), \omega(I_i), \tau(I_{probe}, I_i)\big) = \begin{cases} 1, & \omega(I_{probe}) = \omega(I_i) \text{ and } \tau(I_{probe}, I_i) = 2 \\ 0, & else \end{cases} \tag{16}$$

Recognition rate is computed over different disjoint sets of probe and gallery.

4. Performance Analysis

LGHP has been computed and analyzed on different publically available standard datasets namely Cropped Extended Yale B [23][24], CMU-PIE [25], color-FERET [28][29], and LFW [30]. These databases have been used to test the robustness of the descriptor under severe illumination, pose, expression, and lighting conditions. Performance of LGHP has been compared with state-of-the-art feature descriptors LDP and LVP with respect to APR, ARR and Recognition rate. $L1$ norm distance measure is used as shown in (7) in all the experiments conducted on most challenging facial image databases. Precision Recall space is used to show and compare the Area Under Graph (AUG), as it presents more detailed and clear picture of the performance of the descriptors for highly skewed datasets [26]. To compare the performance of the proposed method different environmental parameters are considered such as pose, illumination, different lighting conditions and expressions. All the facial images have been resized to $64 \times 64$. Histograms of 256 bins are concatenated to generate the descriptors.

Experimental results are computed on a system with Intel(R) Core ™ 2 quad 2.5 GHz CPU, 2GB DDR2 RAM and 32 bit Windows 7 Professional operating system. MATLAB 7.11.0.584 is used to extract and match the features of different datasets.

*4.1    Performance analysis on Extended Yale B database*

Extended Yale B is used in the experiments as it contains images with 64 different illumination variations of 38 subjects. It is a benchmark dataset used to test the robustness of a descriptor against illumination variations. Cropped version of the Extended Yale B dataset is used in the experiments.

Average precision rate (APR) and Average retrieval rate (ARR) of different descriptors for a maximum of 8 retrieved images shown in Fig.5(a) and Fig.5(b). LGHP with $R = 3$ and $R = 2$ shows significant improvement with respect to ARR and APR over LDP and LVP (with $R = 2$). We have shown LVP with $R = 2$ in the results as the performance of LVP is best at $R = 2$. ROC of the proposed descriptor for different R in precision recall space has been shown in Fig.5(c). AUG of the proposed LGHP is small compared to LDP and LVP.

Recognition rates of LGHP, LDP and LVP are computed using (15) and (16) and has been shown in Fig.5(d). Recognition rate of LGHP with $R = 3$ is 84.71% whereas the recognition rates of LVP ($R = 2$) and LDP are 63.67% and 54.88%. Average recognition rates shown in Fig.5(e) are computed by randomly dividing the dataset into disjoint probe and gallery sets of different size. Probe sets are prepared by randomly selecting 20%, 30%, 40%, 50% and 60% images from the datasets and the remaining images are used as corresponding gallery. 10 fold cross validation is used to calculate the average recognition rates for each probe and gallery pair the datasets. Average recognition rate is the average of recognition rates obtained in 10 iterations of the experiment for a particular probe and gallery pair. Recognition rates of the descriptors are computed by modifying the (15) and (16) as follows

$$\gamma = \left(\frac{1}{|PS|}\Sigma_{probe}^{|PS|} \quad \Sigma_{i=1}^{|GS|} \hat{\Delta}\big(\omega(I_{probe}), \omega(I_i), \tau(I_{probe}, I_i)\big)\right) \times 100 \qquad (17)$$

Where $|PS|$ and $|GS|$ are the size of probe and gallery sets.

$\hat{\Delta}(.)$ computes 1st best match for the probe image $I_{probe}$ as

$$\hat{\Delta}\big(\omega(I_{probe}), \omega(I_i), \tau(I_{probe}, I_i)\big) = \begin{cases} 1, & \omega(I_{probe}) = \omega(I_i) \text{ and } \tau(I_{probe}, I_i) = 1 \\ 0, & else \end{cases} \qquad (18)$$

LGHP shows significant improvement even for a very small sized gallery set. Average recognition rates of LGHP($R = 3$), LVP ($R = 2$), and LDP for 60%-40% ratio of the probe and gallery set are 68.31%, 53.72% and 46.16%  respectively.

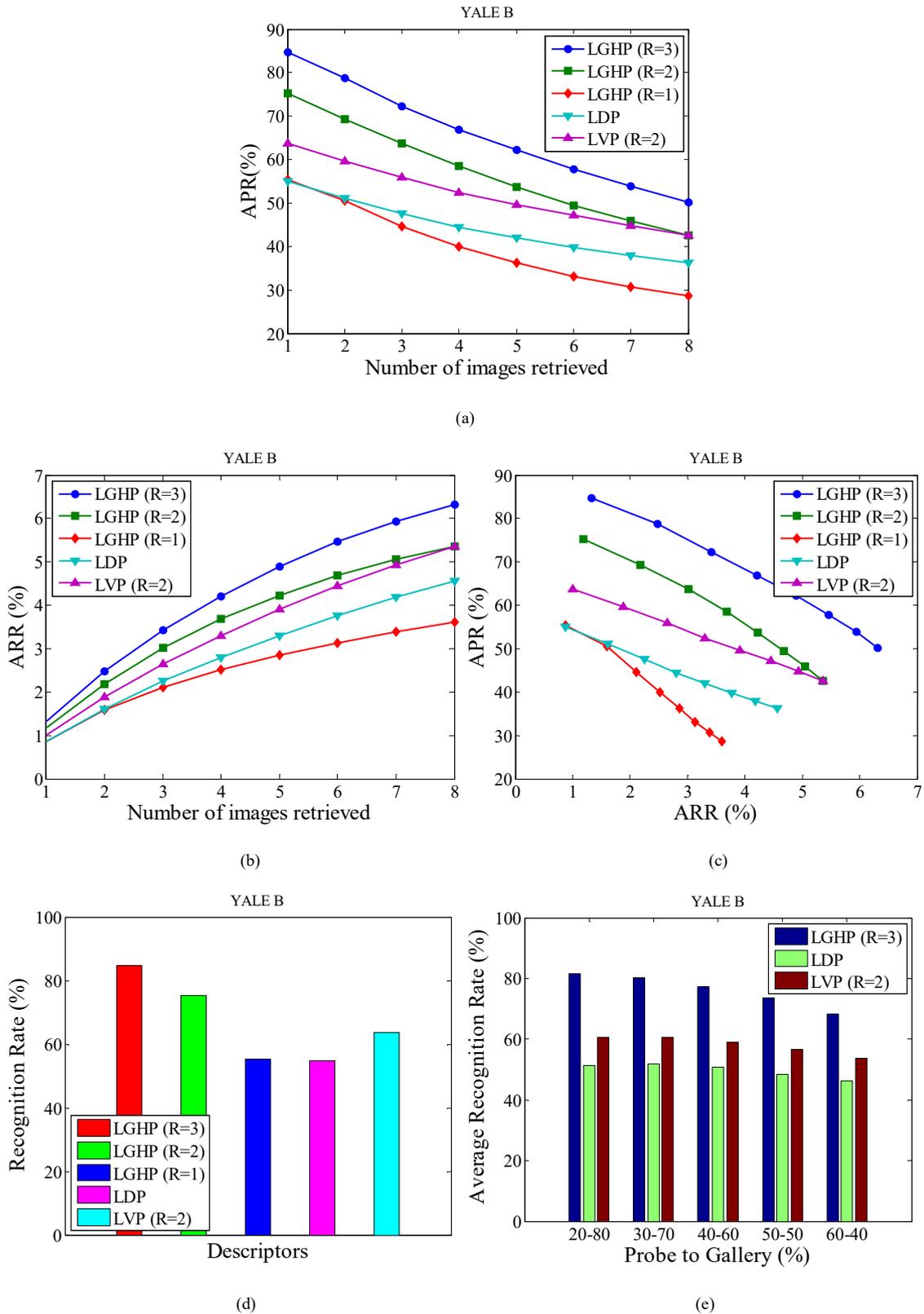

Fig. 5. Comparative results of LGHP, LVP and LDP on Extended Yale B Face datasets (a) APR, (b) ARR, (c) APR vs. ARR, (d) Recognition rates of LDP, LVP ($R = 2$) and LGHP with different values of $R$, and (e) Comparative average recognition rates of LGHP, LVP and LDP of 10-fold cross validation with different size probe and gallery.

*4.2   Performance analysis on CMU-PIE database*

The database contains 38,556 images of 68 subjects taken under varying pose, illuminations, expression and lighting. Expression dataset contains 39 images with different expressions and pose for each subject. Illumination set of the database contains images with 24 different illuminations on 13 different pose of each individual.72 images in 24 different lighting conditions with 3 different pose are in the lights dataset of the database.

APR, ARR and the ROC for different descriptors on expression dataset has been shown in Fig.6(a-c). Results on expression dataset show that LGHP completely outperforms LVP and LDP. LGHP at radii R = 3 and R = 2 performs better than LVP at radii R = 2 and LDP on illumination and lights dataset as shown in Fig.6(d-f) and Fig.6(g-i). We have shown LVP with R = 2 in the results as the performance of LVP is best on CMU-PIE at R = 2. Average precision of LGHP at R = 3, LVP at R = 2, and LDP on expression dataset are 99.59%, 91.58% and 93.83% respectively. Similarly average precision of LGHP at R = 3, LVP at R = 2, and LDP on illumination dataset are 65.97%, 53.68% and 48.18% respectively.

Average recognition rates of descriptors on different datasets shown in Fig.6(j-l) are computed on the same experimental setup as explained in section 4.1. Probe and gallery sets are prepared as elaborated in 4.1. Average recognition rates of LGHP at R = 3, LVP at R = 2, and LDP on expression dataset for 20% probe size are 97.09%, 87.96% and 89.98% respectively. Similarly Average recognition rates of LGHP at R = 3, LVP at R = 2, and LDP on illumination dataset for 20% probe size are 61.97%, 50.19% and 44.97% respectively.

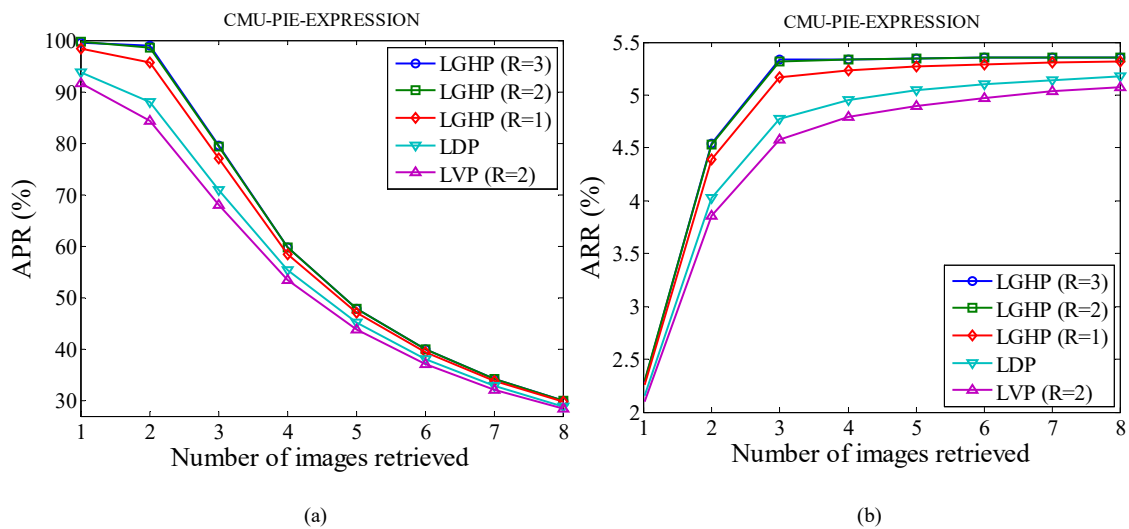

(a)                                          (b)

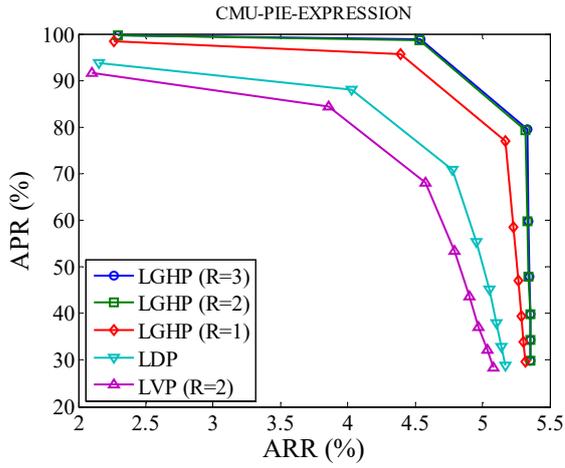

(c)

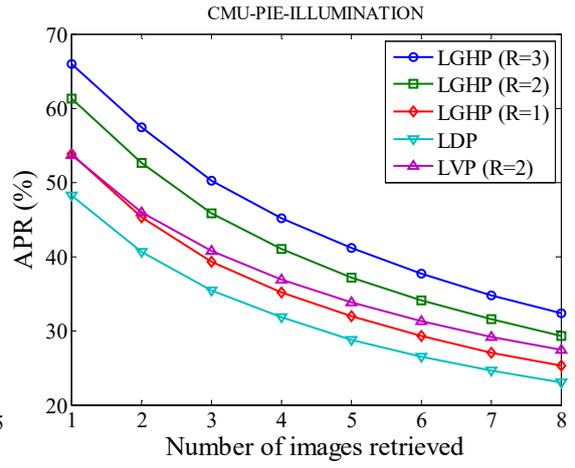

(d)

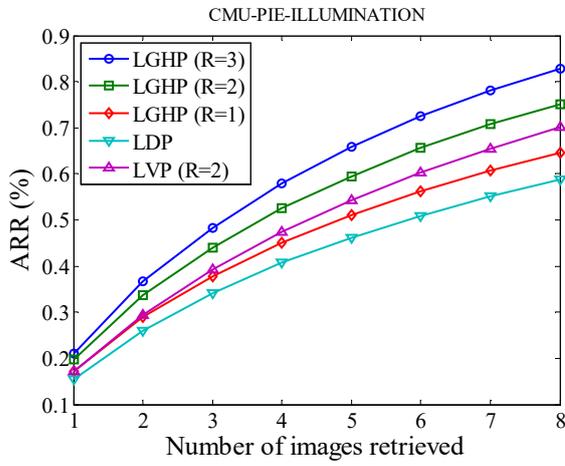

(e)

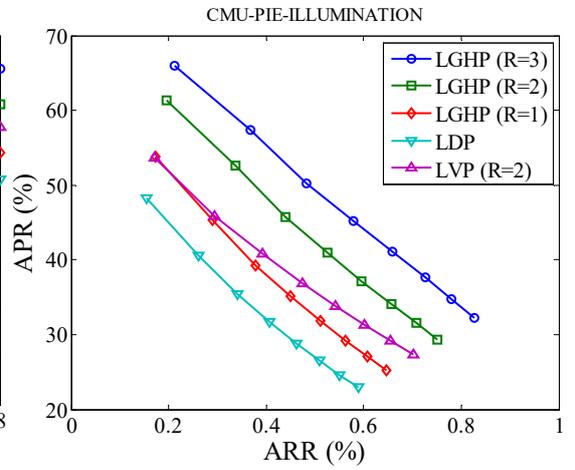

(f)

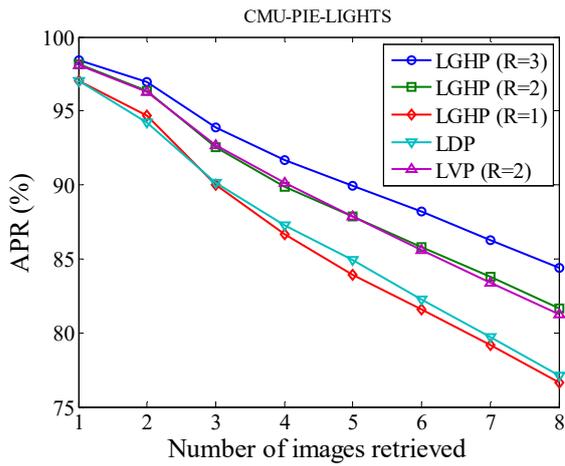

(g)

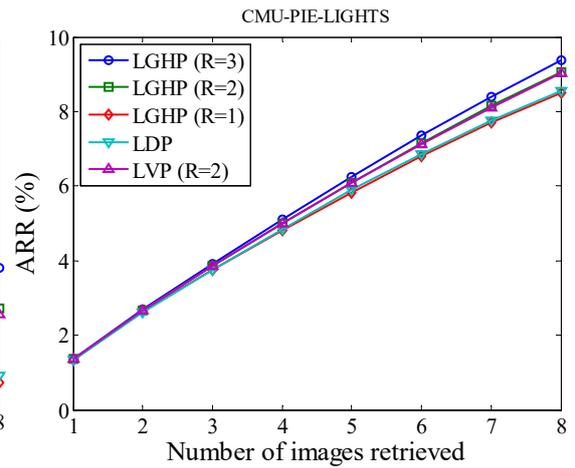

(h)

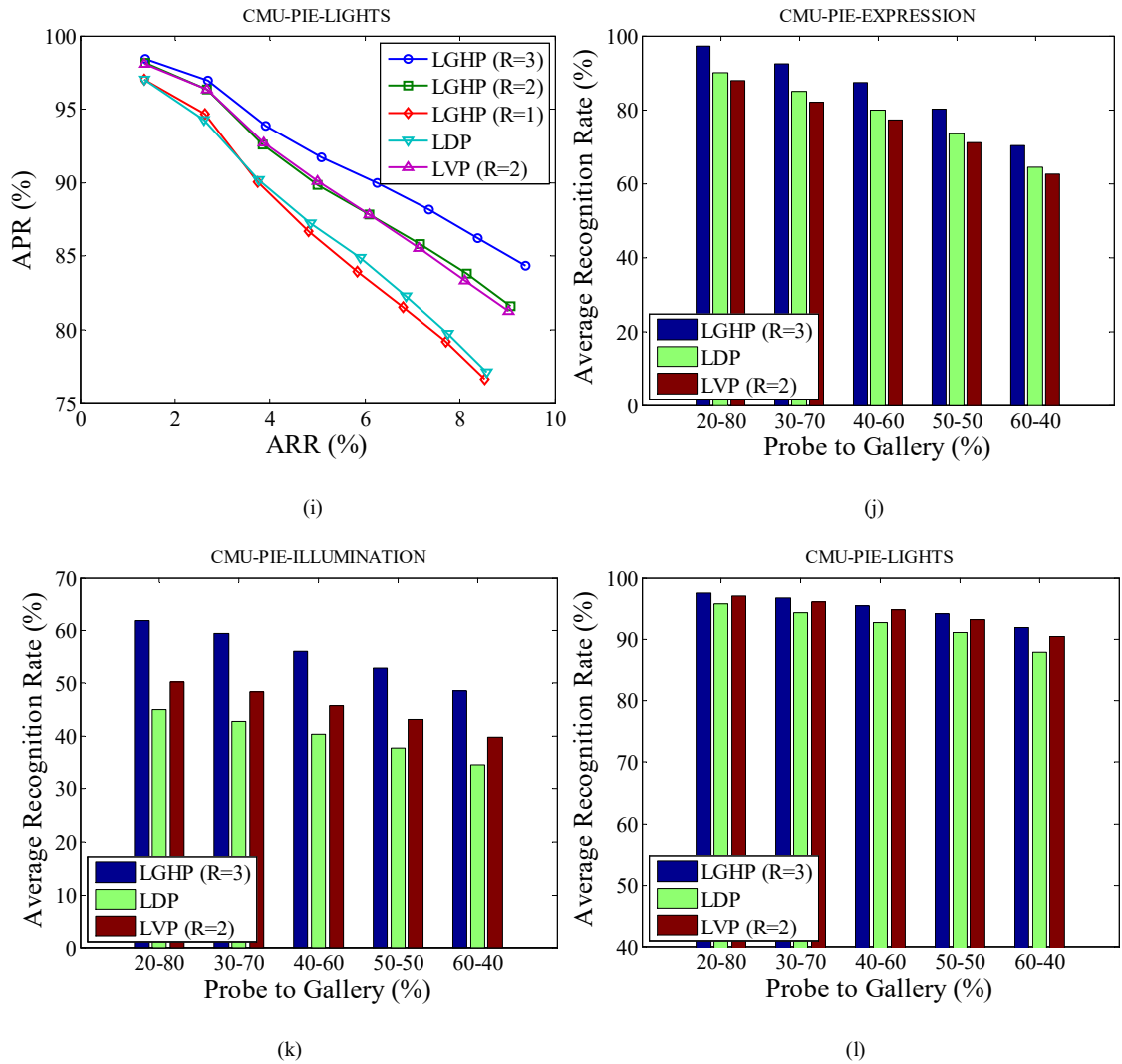

Fig. 6. Comparative results of LGHP, LVP and LDP on CMU-PIE. Results on Expression dataset (a) APR, (b) ARR, (c) APR vs. ARR; Results on Illumination dataset (d) APR, (e) ARR, (f) APR vs. ARR; Results on Lights dataset (g) APR, (h) ARR, (i) APR vs. ARR; Comparative average recognition rates of LGHP, LVP and LDP of 10-fold cross validation with different size probe and gallery (j) Expression dataset, (k) Illumination dataset, and (l) Lights dataset.

### 4.3 Performance analysis on color-FERET database

"Portions of the research in this paper use the FERET database of facial images collected under the FERET program, sponsored by the DOD Counterdrug Technology Development Program Office". Color-FERET database is one of the most challenging facial image databases with severe variations in pose and expression. The color FERET database contains 11,338 facial images of 994 individuals at different angles. There are 13 different pose used in the images of the database [28][29]. Brief description of pose has been given in Table1. The gallery used in the experiments contains images of those individuals having 20 or more images in the database. APR and ARR shown for different descriptors along with their minimized version (computed using uniform 2 pattern minimization). Minimized versions of LGHP, LVP, and LDP are named as LGHPU2, LVPU2, and LDPU2 respectively. Uniform 2 (U2) patterns are those binary patterns which have at most two transitions from 0 to 1 or vice versa [16].

APR and ARR shown in Fig.7(a) and Fig.7(b) illustrates that LGHP at different radii completely outperforms LDP and LVP. Even the minimized version of LGHP performs better than LVP at $R = 3$, LVPU2 at $R = 3$, and LDPU2. Maximum APR of LGHP is almost 90% at $R = 3\ and\ R = 2$. The APR of LVP at $R = 2$, which is nearest to the maximum APR of LGHP is less than 80%. Maximum ARR of LGHP is more than 15% which is at least 3% more than the ARR of its closest counterpart.

AUG for LGHP at $R = 3\ and\ R = 2$ shown in Fig.7(c) is less than the AUG of all other descriptors and their variations. Recognition rates shown in Fig.7(d) are computed using experimental settings as elaborated in section 4.1. Maximum recognition rate of LGHP is more than 80%, which is at least 7% more than the recognition rate of its nearest counterpart. Hence the proposed descriptor shows significant improvement in recognition and retrieval rates compared to LDP and LVP and their other variations.

Table 1: Description of different pose in images of color-FERET database.

| Pose Name | Description |
| --- | --- |
| fa | regular frontal image |
| fb | alternative frontal image, taken shortly after the corresponding fa image |
| pl | profile left |
| hl | half left - head turned about 67.5 degrees left |
| ql | quarter left - head turned about 22.5 degrees left |
| pr | profile right |
| hr | half right - head turned about 67.5 degrees right |
| qr | quarter right - head turned about 22.5 degrees right |
| ra | random image - head turned about 45 degree left |
| rb | random image - head turned about 15 degree left |
| rc | random image - head turned about 15 degree right |
| rd | random image - head turned about 45 degree right |
| re | random image - head turned about 75 degree right |

*4.4    Performance analysis on LFW database*

There are 13,233 color facial images of 5,749 individuals. 1680 individuals have two or more images and rest of the individuals have only one image. As the problem of unconstrained face recognition is one of the most general and fundamental face recognition problems we test the proposed descriptors on the most challenging facial image database LFW. Experimental analysis is done on a subset of LFW database. In the experiments images of those individuals having at least 20 images are taken. Similar settings are used for experiments as used in section 4.3. APR, ARR, and recognition rates for LGHP, LVP, LDP and their U2 versions have been shown in Fig.8. Maximum APR of LGHP is approximately 2% more than the maximum APR of its immediate counterpart whereas maximum ARR is approximately 0.5% more than the maximum ARR of its immediate counterpart.

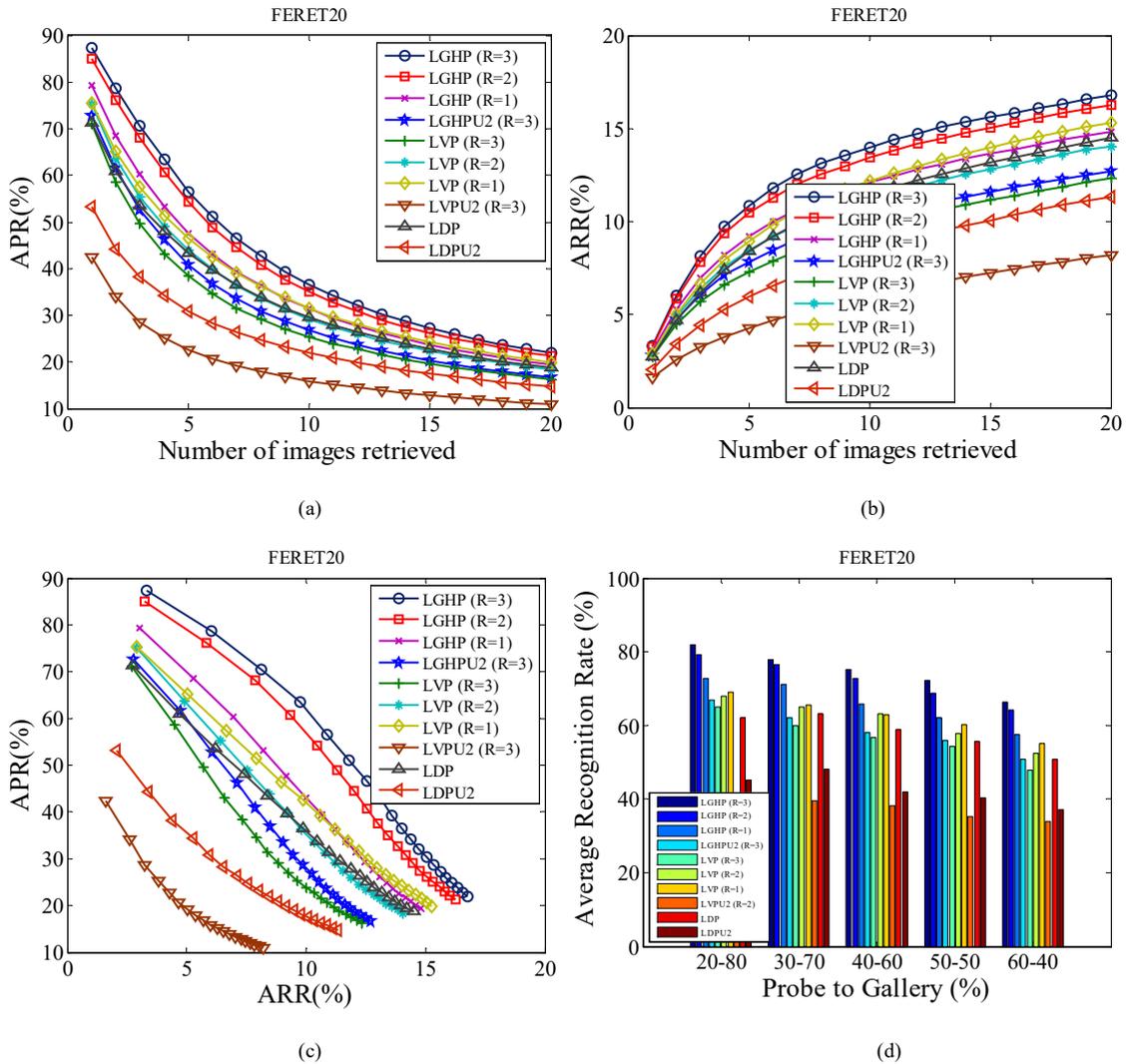

Fig. 7. Comparative results of LGHP, LVP and LDP on FERET Face datasets along with uniform 2 minimization (a) APR, (b) ARR, (c) APR vs. ARR, (d) Comparative average recognition rates of LGHP, LVP and LDP of 10-fold cross validation with different size probe and gallery.

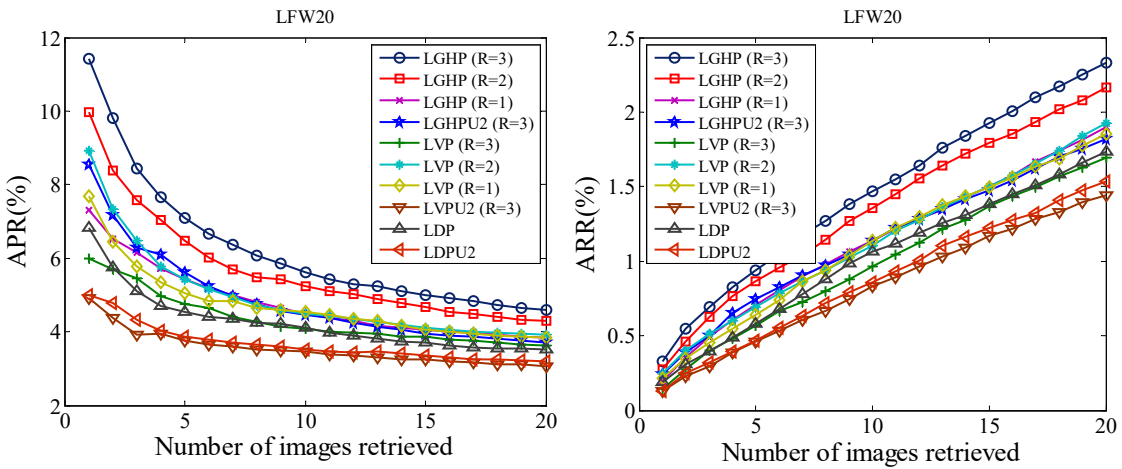

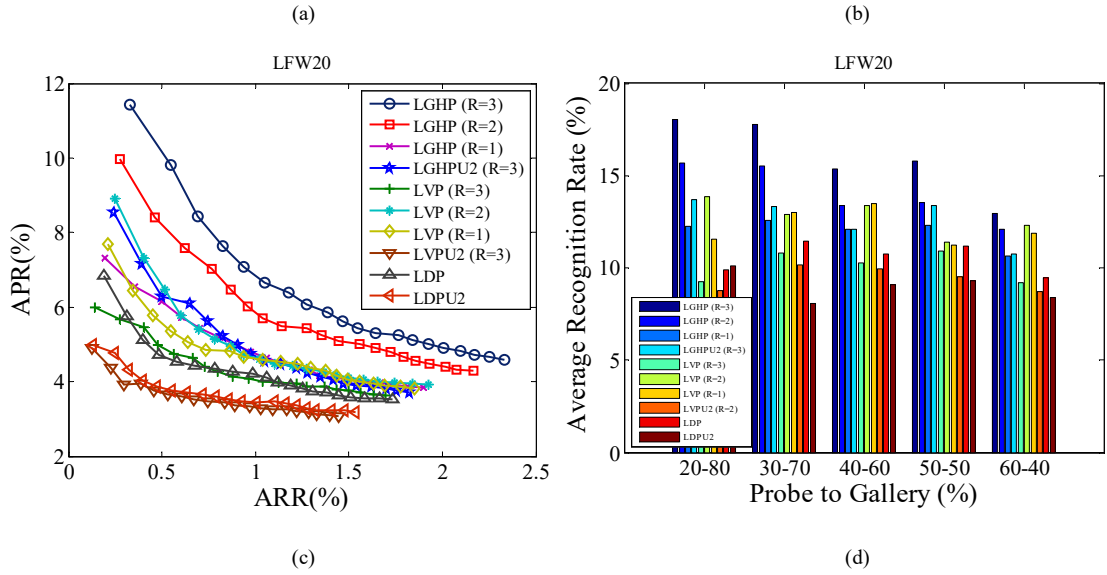

Fig. 8. Comparative results of LGHP, LVP and LDP on LFW Face datasets along with uniform 2 minimization (a) APR, (b) ARR, (c) APR vs. ARR, (d) Comparative average recognition rates of LGHP, LVP and LDP of 10-fold cross validation with different size probe and gallery.

AUG shown in Fig.8(c) demonstrates that the proposed descriptor maintains its superiority over other related descriptors with increasing retrieval (recall) rates. LGHP performs better than most of the other descriptors and their variations even when U2 minimization has been used. Maximum recognition rate of LGHP shown in Fig.8(d) is 4% more than the maximum recognition rate achieved by other descriptors.

A 2-dimensional Gabor filter [22] shown in (19) has been used to compute the Gabor response with two scales and two directions (0°, 90°). A Gabor filter is a Gaussian response over sinusoid with frequency $f$ and standard deviations $\sigma_s$ and $\sigma_t$ [22].

$$\emptyset(s,t) = \frac{1}{2\pi\sigma_s\sigma_t} e^{[-(1/2)(s^2/\sigma_s^2 + t^2/\sigma_t^2) + 2\pi i f s]} \tag{19}$$

Gabor features are computed using different descriptors over Gabor response obtained by convolving Gabor filter with the original image.

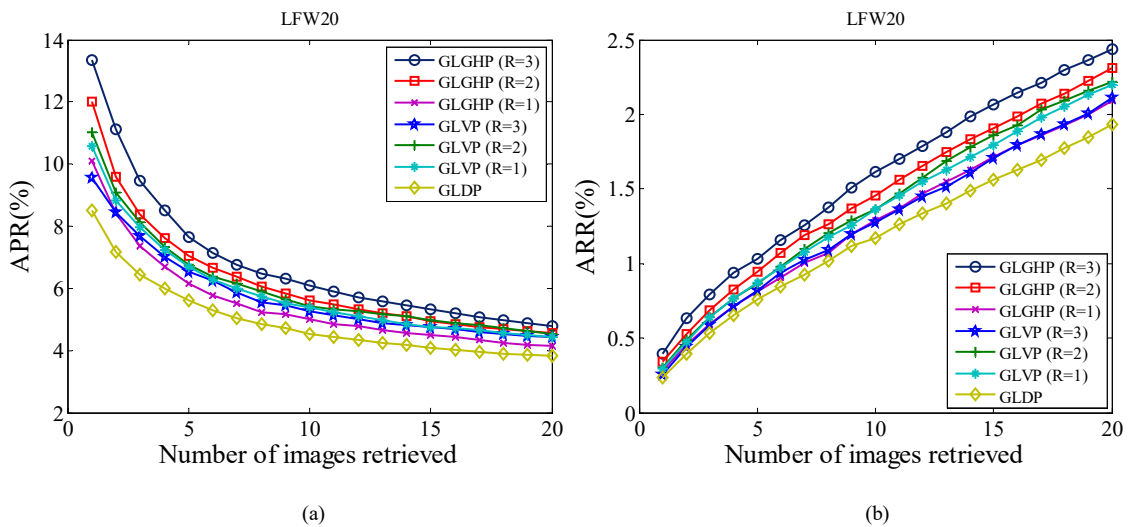

(a)     (b)

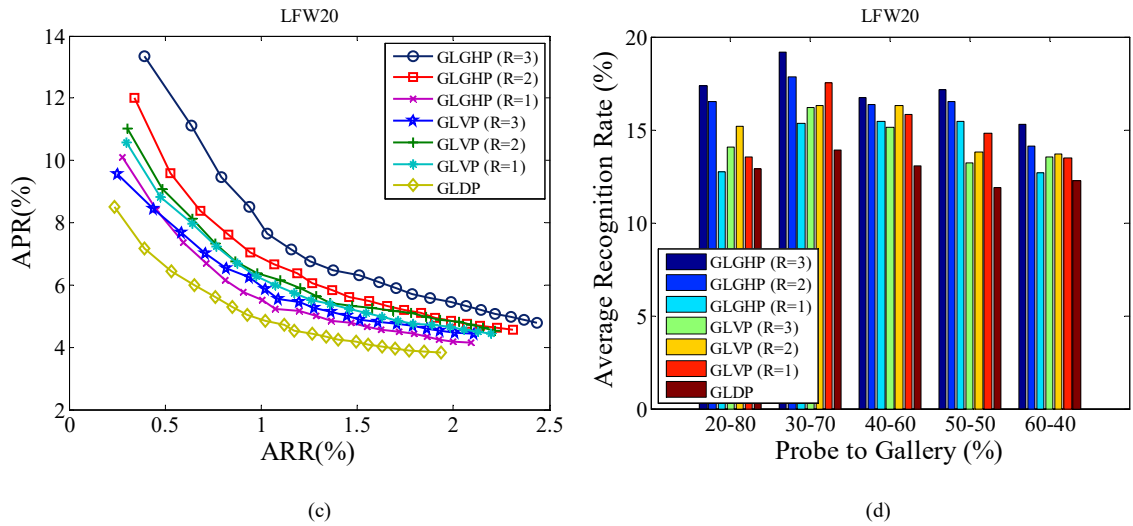

(c)                                (d)

Fig. 9. Comparative results of Gabor features GLGHP, GLVP and GLDP corresponding to LGHP, LVP, and LDP on LFW Face datasets (a) APR, (b) ARR, (c) APR vs. ARR, (d) Comparative average recognition rates of GLGHP, GLVP and GLDP of 10-fold cross validation with different size probe and gallery.

APR, ARR, and recognition rates of Gabor features corresponding to different descriptors namely GLGHP (corresponding to LGHP), GLVP (corresponding to LGVP), and GLDP (corresponding to LGVP) are shown in Fig.9. Gabor features improves the performance of all the descriptors even more so the proposed descriptor LGHP. Maximum APR of GLGHP shown in Fig.9(a) is approximately 14%, which is 4% more than the APR of the immediate best counterpart. The maximum recognition rate of GLGHP shown in Fig.9(d) is approximately 20%, which is 2% more than the maximum recognition rate of GLVP.

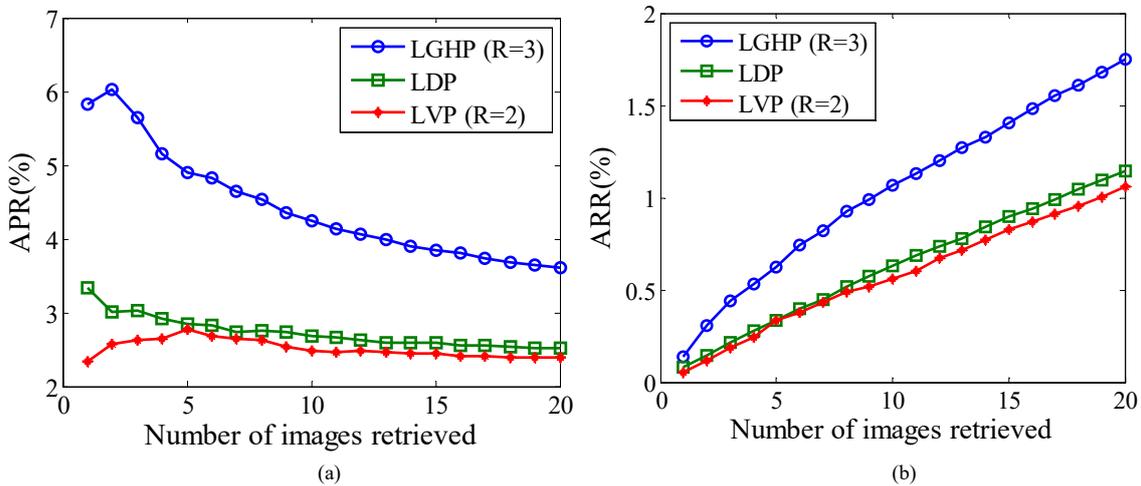

(a)                                (b)

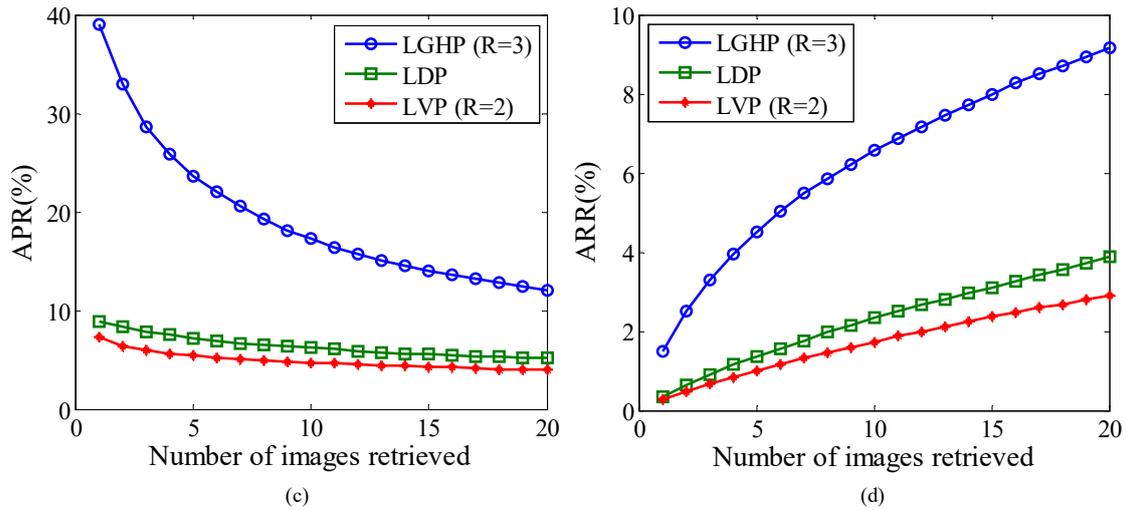

Fig. 10. Comparative results of the descriptors under Additive White Gaussian Noise (AWGN) with zero mean and 0.05 variance: (a) APR on LFW, (b) ARR on LFW, (c) APR on FERET, and (d) ARR on FERET.

*4.5  Performance analysis under noise on LFW and FERET databases*

Performance of the proposed descriptor has been analyzed and compared with LDP and LVP under Additive White Gaussian Noise (AWGN) with zero mean and 0.05 variance. Features of the facial images in LFW and FERET databases are computed after adding the AWGN. Feature of each noisy image taken as the query. APR and ARR values are computed for each queried feature. Computed APR and ARR values are shown in Fig. 10. Fig. 10(a) and (b) show the results computed on LFW. Proposed descriptor shows approximately 1.5% improvement in APR and ARR over its nearest counterpart LDP under noise. Results computed over FERET database under noise are shown in Fig. 10(c) and (d). LGHP shows significant improvement of about 10% in APR and 7% in ARR as shown in Fig. 10(c) and (d). As the results show the proposed descriptor is robust enough to achieve better retrieval accuracy under noisy conditions.

## 5. Conclusion

In this paper a local descriptor has been proposed for face recognition and facial image retrieval. Performance of the proposed descriptor has been tested on benchmark databases and compared with the state-of-the-art descriptors to show the robustness of the proposed descriptor against pose, expression, illumination, and lighting variations. Experimental results demonstrate the superior performance over the existing state of the art descriptors under huge variations in pose, expression, lighting and illumination conditions. The proposed descriptor has the potential to be used in the real world unconstrained facial recognition and retrieval application.